\documentclass{article}

\usepackage{arxiv}

\usepackage[utf8]{inputenc} 
\usepackage[T1]{fontenc}    
\usepackage{hyperref}       
\usepackage{url}            
\usepackage{booktabs}       
\usepackage{amsfonts}       
\usepackage{nicefrac}       
\usepackage{microtype}      
\usepackage{lipsum}
\usepackage{graphicx}
\graphicspath{ {./images/} }

\title{Review of Digital Asset Development with Graph Neural Network Unlearning}

\author{
 Zara Lisbon \\
  Independent Researcher\\
}

\begin{document}
\maketitle
\begin{abstract}
In the rapidly evolving landscape of digital assets, the imperative for robust data privacy and compliance with regulatory frameworks has intensified. This paper investigates the critical role of Graph Neural Networks (GNNs) in the management of digital assets and introduces innovative unlearning techniques specifically tailored to GNN architectures. We categorize unlearning strategies into two primary classes: data-driven approximation, which manipulates the graph structure to isolate and remove the influence of specific nodes, and model-driven approximation, which modifies the internal parameters and architecture of the GNN itself. By examining recent advancements in these unlearning methodologies, we highlight their applicability in various use cases, including fraud detection, risk assessment, token relationship prediction, and decentralized governance. We discuss the challenges inherent in balancing model performance with the requirements for data unlearning, particularly in the context of real-time financial applications. Furthermore, we propose a hybrid approach that combines the strengths of both unlearning strategies to enhance the efficiency and effectiveness of GNNs in digital asset ecosystems. Ultimately, this paper aims to provide a comprehensive framework for understanding and implementing GNN unlearning techniques, paving the way for secure and compliant deployment of machine learning in the digital asset domain.
\end{abstract}


\section{Introduction}
The rise of digital assets, including cryptocurrencies, decentralized finance (DeFi) tokens, and non-fungible tokens (NFTs), has revolutionized the global financial landscape. Digital assets operate on decentralized networks, primarily using blockchain technology, which enables secure, transparent, and immutable record-keeping. However, as the use of blockchain and digital assets grows, so do concerns related to data privacy, security, and compliance with regulatory frameworks like the General Data Protection Regulation (GDPR). These regulations often require users to have control over their data, including the ability to have it removed or forgotten—a notion seemingly at odds with blockchain's immutable nature.

One of the emerging approaches to addressing these challenges is Graph Neural Network (GNN) unlearning, a method designed to enable machine learning models, particularly GNNs, to "forget" specific data points after they have been incorporated into a trained model. GNNs are particularly relevant in the digital asset space due to their capacity to model complex relationships in blockchain transaction networks, which often take the form of graphs. This paper explores the concept of Graph Neural Network Unlearning for digital assets, focusing on the importance of privacy, potential use cases, and methods to implement effective unlearning without compromising the integrity and utility of GNNs.

\section{Literature Review}
There are two main types of unlearning techniques in the context of machine learning models, including Graph Neural Networks (GNNs):
1. Exact Unlearning and 2. Approximate Unlearning. According to \cite{lina2024mutaxonomy}, the basic goal for exact unlearning is to align the unlearned model's distribution with that of a restrained model, while approximate unlearning has indistinguishable parameters between the unlearned and retrained model.
\par Exact unlearning involves completely removing the influence of specific data points from a trained machine learning model. This is typically done by retraining the model from scratch without the data that needs to be forgotten. In the case of GNNs, this would require retraining the entire graph model after removing the nodes or edges corresponding to the data that should be unlearned (\cite{kipf2016semi}).
The model is guaranteed to no longer retain any information about the removed data, providing strong privacy guarantees. The retrained model is as if the deleted data was never included in the first place. The drawbacks of exact unlearning is that retraining can be computationally expensive, especially for large graphs in digital asset applications (e.g., cryptocurrency transaction networks or DeFi systems). This method also requires significant resources and time, making it impractical for real-time or large-scale systems.

\par Approximate unlearning aims to minimize the impact of specific data points without completely retraining the model. Instead of starting over, the model is incrementally updated to “forget” the data in question. This can be achieved by reversing the gradients of the data that need to be forgotten or selectively updating the affected parts of the model. Approximate unlearning is far more efficient and scalable than exact unlearning. It saves computational resources and time, making it a more feasible solution for real-time applications and large-scale systems like digital asset networks. However, this method may leave behind residual information about the deleted data, raising concerns about privacy in highly sensitive applications. The model might not completely "forget" the data, which could be problematic for meeting strict privacy regulations such as GDPR.

\par GNN was originally focusing on image and tabular data (\cite{lina2024mutaxonomy}). \cite{chen2022graph} proposes the first GNN exact unlearning approach called GraphEraser. GraphEraser introduces two balanced partition methods to retain graph structural information. Later, \cite{wang2023inductive} created GUIDE that ensures fairness and  balance constraints in graph partitioning. GNN has many mature application in representing complex structure like in \cite{wang2024graph}.
\section{Exact Unlearning for GNN}
Exact unlearning utilize a machine learning model in which individual components are trained on disjoint subsets of the data. During deletion, exact unlearning approaches only retrain the affected components rather than the entire model. \cite{lina2024mutaxonomy} classifies exact unlearning with conventional model with convex function and complex model with non-convex function. Convex function models are like those introduced in \cite{jose2021unified} and \cite{kashef2021boosted}. Complex models are like optimization problem in \cite{elhedhli2017airfreight}, where non-convex may produce local optimal solutions, that leads to more computation resourced required than convex optimization. GNN comes into the picture to help with those non-convex structure. 

\par For GNNs, exact unlearning requires retraining the entire graph model from the beginning without the specific nodes, edges, or subgraphs that were originally included. This guarantees that the model behaves as if the data had never been part of the training process.
\section{Approximate Unlearning for GNN}
\cite{lina2024mutaxonomy} and \cite{shaik2023exploring} categorize approximate unlearning into two classes: data-driven approximation and model-driven approximation. Strategies that focus on manipulating the data are categorized as data driven approximation.  \cite{he2021deepobliviate}, \cite{gupta2021adaptive} and \cite{neel2021descent} use data isolation strategy for data-driven approximate unlearning. Model-driven approximation, on the other hand, focuses on the internal components of the model itself rather than the data. This approach modifies the learned parameters of the model or its architecture to approximate the removal of data influence. Both approaches aim to remove the influence of specific data points from machine learning models, but they differ in their methodologies and the specific components of the learning system they target. 

\subsection{Data-Driven Approximation in GNNs}
Data-driven approximation focuses on manipulating the training data itself to achieve unlearning. In the context of GNNs, this can involve techniques that isolate or modify the graph structure to remove specific nodes (representing data points) and their corresponding edges (representing relationships or interactions). As noted in works like \cite{he2021deepobliviate}, \cite{gupta2021adaptive}, and \cite{neel2021descent}, data isolation can be implemented in GNNs by selectively removing nodes from the graph. This means that when a specific user or transaction needs to be unlearned, the corresponding node and its connections can be removed without affecting the remaining graph.
\paragraph{Data Isolation Strategies}
The GNN can then be retrained or updated to reflect this modification, ensuring that the influence of the unlearned data is effectively eradicated. However, this approach may incur additional computational costs, particularly when dealing with large graphs, since it requires a thorough update of the affected areas.
\paragraph{Graph Modification Techniques} Techniques that alter the graph structure directly, such as edge deletion or node merging, fall under the data-driven category. By modifying the edges connecting the nodes, the relationships can be redefined, effectively 'forgetting' the influence of the specific data points on the GNN’s predictions.
\paragraph{Challenges and Trade-offs} The main challenge with data-driven approaches in GNNs is maintaining model utility while removing specific data influences. GNNs are inherently sensitive to their structure; thus, significant alterations to the graph can potentially degrade the model's performance on tasks such as node classification or link prediction.
\subsection{Model-Driven Approximation in GNNs}
Model-driven approximation, on the other hand, centers around modifying the model's internal components rather than the data itself. In the context of GNNs, this could involve adjusting the learned parameters or architecture of the model to achieve the desired unlearning effects.

\paragraph{Parameter Adjustment}
GNNs utilize learned parameters (weights) during the aggregation and transformation of node features. Model-driven unlearning could involve recalibrating these parameters in a way that diminishes the influence of specific nodes without needing to physically remove them from the graph.
Techniques such as inverse gradient descent or parameter perturbation can be applied to adjust the weights associated with unlearned nodes, effectively reducing their impact on the model’s predictions without retraining from scratch.
\paragraph{Architecture Modification}
Changes to the GNN architecture, such as altering the number of layers or the aggregation functions, can also represent a form of model-driven approximation. By tweaking how information is processed, the model can reduce the influence of specific data points. For instance, if certain layers are found to overemphasize the features of unlearned nodes, these can be adjusted or restructured to mitigate their effect.
\paragraph{Efficiency and Scalability}
Model-driven approaches can be more computationally efficient than data-driven methods because they avoid the overhead of reconstructing the graph. Instead of re-training on a modified dataset, model-driven strategies focus on adjusting the existing model parameters, which can often be achieved with lower resource consumption.
\subsection{Comparative Analysis and Relationship to GNNs}
The primary difference between data-driven and model-driven approximations lies in their focus: data-driven approaches emphasize modifications to the dataset and its structure, while model-driven strategies concentrate on the internal mechanics of the model.

\paragraph{Influence on Performance} In GNNs, data-driven unlearning may lead to performance degradation due to the loss of structural information inherent in the graph. In contrast, model-driven unlearning might preserve more of the graph's relational properties while achieving the same objective.

\paragraph{Suitability for Different Applications} The choice between these methods often depends on the application. For instance, in contexts requiring stringent privacy compliance (like financial transactions), data-driven unlearning may be preferred for its directness. In scenarios where computational efficiency is paramount (such as real-time risk assessment), model-driven techniques may be more suitable.

\paragraph{Complementary Approaches} Both strategies can be viewed as complementary. Hybrid approaches that incorporate aspects of both data-driven and model-driven unlearning may provide the best outcomes, allowing for a balance between structural integrity, computational efficiency, and compliance with unlearning requirements.

\section{GNN Unlearning for Digital Assets}
\subsubsection{Motivation and Importance}
The core motivation behind GNN unlearning in the context of digital assets stems from the need to balance transparency and immutability with data privacy and security \cite{shokri2015privacy}. While blockchain transactions are inherently designed to be permanent and publicly accessible, personal and sensitive data linked to these transactions might need to be erased or forgotten in some instances. For example, if a user’s wallet data or a transaction that contains sensitive information is accidentally included in a publicly accessible GNN-based fraud detection model, the user might have the right to request its removal under various privacy regulations. Achieving this in a decentralized and immutable environment is challenging, especially when the data is integrated into machine learning models such as GNNs.

\par Additionally, digital assets often involve highly dynamic graphs. A cryptocurrency transaction network, for instance, can expand and contract in real-time as new transactions and wallets are created, and as older transactions lose relevance. Unlearning specific nodes (e.g., user wallets or transactions) or edges (e.g., relationships between wallets) from a GNN model trained on such a network without retraining the entire model is critical for maintaining operational efficiency. The ability to selectively unlearn data ensures that digital asset models stay compliant with evolving regulatory environments while continuing to provide accurate and real-time predictions, such as risk assessments or market forecasts.
\par The Role of GNNs in Digital Asset Applications
Graph Neural Networks have become a powerful tool for understanding and predicting patterns in blockchain networks. A GNN can model the interactions between wallets, transactions, smart contracts, and digital asset exchanges, allowing for a deeper understanding of the financial ecosystem. For instance, GNNs are used for fraud detection in cryptocurrency transactions by identifying suspicious patterns of behavior across related entities. Similarly, GNNs can analyze the flow of tokens within decentralized finance applications to provide better credit scoring and assess risks in lending markets.
\par In these applications, nodes represent entities such as wallets, transactions, or contracts, while edges represent the relationships between them, such as token exchanges, transaction histories, or trust ratings. This graph structure is well-suited for detecting patterns that would be difficult to recognize using traditional machine learning models. However, the very nature of this data poses unique challenges when it comes to compliance with privacy regulations. For example, if a GNN model has learned from a set of transactions that involve a user who later requests their data to be erased, simply deleting the user’s data from the model is insufficient due to the complex relational dependencies that GNNs leverage.

\subsection{Core Challenges in GNN Unlearning for Digital Assets}
Unlearning in GNNs, particularly for digital asset applications, faces unique challenges that make the process more complex compared to traditional unlearning tasks. These challenges stem from both the structure of blockchain networks and the real-time, dynamic nature of digital asset markets.

\paragraph{Immutability and Transparency} Blockchain’s core feature of immutability ensures that all transactions are permanent and verifiable by anyone. This presents a fundamental conflict with the concept of "forgetting" or unlearning. In a GNN that models these transactions, the historical data used to train the model is theoretically available indefinitely, even if it has been unlearned from the model. Achieving a balance between privacy and blockchain’s transparency is a critical challenge.

\paragraph{Relational Data and Dependencies} GNNs excel at capturing the relationships between entities in a graph, which makes unlearning specific nodes or edges complex. In digital asset applications, unlearning a wallet or transaction affects not only the specific node but also its neighbors and the larger network. The removal of one node might disrupt the GNN’s understanding of relationships between other nodes, potentially leading to a loss in model accuracy and utility.

\paragraph{Dynamic Graph Structures} Cryptocurrency networks and DeFi applications are highly dynamic, with transactions and nodes being added or removed continuously. In such a fast-evolving environment, it is necessary to design GNN unlearning mechanisms that can adapt to the constantly changing graph structure without requiring complete retraining, which is computationally expensive and inefficient for real-time applications.

\paragraph{Adversarial Risks} In the context of digital assets, adversarial attacks aimed at reconstructing deleted or unlearned data pose a significant threat. Even after removing a node or edge from a GNN, there may be residual information that can be exploited by an attacker to infer the deleted data. Ensuring that GNN unlearning methods provide strong security guarantees is essential for preserving user privacy.

\paragraph{Efficiency and Scalability} Blockchain networks can consist of millions of nodes and edges, especially when considering transaction histories spanning years or more. Efficient unlearning techniques that scale to such large graphs are crucial. Traditional methods, like retraining from scratch, are impractical due to the computational resources and time required, emphasizing the need for innovative, scalable unlearning solutions.

\subsection{Methods for GNN Unlearning in Digital Assets}
Addressing the challenges of GNN unlearning for digital asset applications requires a range of techniques that balance efficiency, scalability, and privacy. Below are some of the potential methods for achieving effective unlearning in this domain:

\paragraph{Exact Unlearning via Retraining} The most straightforward unlearning approach is to retrain the GNN from scratch after removing the specific data points. This method ensures that the model no longer retains any traces of the deleted nodes or edges. However, retraining is highly computationally expensive, especially in blockchain networks where the graph may involve millions of nodes and transactions. While this guarantees that the data is forgotten, the method is impractical for most real-time digital asset applications.

\paragraph{Approximate Unlearning} Approximate unlearning methods aim to minimize the computational overhead by updating only the portions of the GNN affected by the deleted data. This can be done by reversing the gradients associated with the removed nodes or edges, effectively "undoing" their impact on the model without retraining the entire graph. Although faster and more scalable, approximate unlearning methods may leave residual information, which could be problematic in privacy-sensitive applications like digital asset management.

\paragraph{Subgraph-Based Unlearning} In many blockchain-based systems, only a localized subgraph (e.g., a group of related transactions or wallets) is affected by the removal of a specific entity. Subgraph-based unlearning techniques focus on isolating the affected portion of the graph and retraining only that subgraph, leaving the rest of the model intact. This approach can dramatically reduce the computational cost of unlearning, making it more suitable for large-scale blockchain networks.

\paragraph{Embedding-Based Unlearning} GNNs rely heavily on node embeddings to represent entities in the graph. Embedding-based unlearning methods focus on erasing or modifying the embeddings associated with the deleted data. This approach is computationally efficient and particularly well-suited for large graphs, as it avoids retraining the entire model. However, it requires careful attention to ensure that the modified embeddings no longer contain information about the removed data while preserving the model’s overall utility.

\paragraph{Differential Privacy and Adversarial Regularization} In highly sensitive applications, ensuring that the unlearned data cannot be reconstructed is of utmost importance. Differential privacy can be applied to GNNs by adding noise to the model’s parameters, ensuring that individual data points cannot be inferred from the model’s outputs. Additionally, adversarial regularization can be used to penalize the model for retaining information about unlearned data, providing further security guarantees.

\subsection{Use Cases for GNN Unlearning in Digital Assets}
Graph Neural Network unlearning for digital assets has several key use cases, each with its own set of privacy and security requirements. Below, we explore some of the most important applications of GNN unlearning in this domain:

\paragraph{Fraud Detection in Cryptocurrencies} GNNs are commonly used to detect fraudulent activities in cryptocurrency networks by analyzing transaction patterns and relationships between wallets. If a legitimate user is mistakenly flagged as fraudulent and requests their data be erased, GNN unlearning techniques can be employed to remove the user’s transaction history from the model without affecting the detection of other fraudulent activities. This ensures that the model remains accurate while respecting the user’s privacy.

\paragraph{NFT Recommender Systems} Non-fungible tokens (NFTs) rely on blockchain networks to track ownership and transactions. GNNs can be used to build recommender systems that suggest NFTs based on users’ past interactions. If a user requests that their NFT activity be forgotten, GNN unlearning methods can be applied to remove the user’s transaction history from the recommendation model. This is particularly relevant for NFT marketplaces that must comply with privacy regulations while still providing personalized recommendations.
\paragraph{Decentralized Finance (DeFi) Risk Assessment}
Decentralized Finance (DeFi) applications leverage blockchain networks to facilitate lending, borrowing, and trading without intermediaries. GNNs play a crucial role in assessing risk by analyzing the relationships between users, smart contracts, and transaction histories (\cite{schar2021decentralized}). In cases where users or contracts need to be unlearned (for example, if a user requests data removal under GDPR), GNN unlearning ensures that the data is erased without negatively impacting the DeFi application’s ability to evaluate ongoing risks. Unlearning the data of a particular contract or wallet could be critical when users revoke permissions for lending agreements, or when smart contracts are deemed obsolete or flawed. The ability to maintain the accuracy of risk assessments while complying with user privacy requirements becomes essential in this fast-evolving financial ecosystem.

\paragraph{Token Relationship Prediction}
Another important use case for GNN unlearning in digital assets is predicting relationships between different tokens, such as in decentralized exchanges (DEXs) where users trade assets across various blockchain networks. GNNs model the interactions between these tokens by analyzing historical trade data, liquidity pools, and market conditions. When a user, token issuer, or exchange requests the removal of data related to a particular token or trade, GNN unlearning enables the model to "forget" those specific transactions while preserving its ability to predict new token relationships accurately (\cite{zhang2018link}). This is particularly useful for regulatory compliance in jurisdictions where certain tokens may become restricted or blacklisted, necessitating the removal of their associated trade histories.

\paragraph{Blockchain-Based Identity Management}
With the rise of self-sovereign identity (SSI) and decentralized identity systems on blockchain, GNNs are often used to manage and authenticate relationships between users, organizations, and credentials. These identities are represented as nodes in a graph, and the edges capture various interactions, such as verification processes and access permissions. If a user requests to remove their identity from the system, GNN unlearning can selectively erase their node and related credentials from the model. This enables compliance with privacy regulations without disrupting the overall system, ensuring that other users and relationships remain intact. As identity management becomes a cornerstone of decentralized networks, the ability to remove identities without retraining the entire graph is essential for scalability and privacy.

\paragraph{Blockchain Governance and Voting Systems}
Decentralized governance mechanisms, such as those used in decentralized autonomous organizations (DAOs), rely on blockchain to facilitate voting, proposal submissions, and consensus decisions. GNNs can be employed to analyze voting patterns, predict governance outcomes, and model relationships between participants in governance decisions. If a participant in the governance process wishes to withdraw their voting history or erase their participation data, GNN unlearning ensures that this information can be removed without undermining the governance structure. This capability is especially critical when handling sensitive governance matters, where participants may want to disassociate themselves from certain decisions while maintaining the integrity of the governance model.

\paragraph{Regulatory Compliance in Financial Markets}
As governments and regulatory bodies increase oversight of digital assets, organizations must adapt to regulations that require the deletion or obfuscation of certain data. For instance, anti-money laundering (AML) and know-your-customer (KYC) regulations in cryptocurrency markets often necessitate the collection and storage of user data for compliance. However, when a user’s data must be deleted due to a change in legal status or a regulatory directive, GNN unlearning can remove the user’s transaction data, while still preserving the rest of the graph for ongoing compliance activities. This ensures that financial institutions and exchanges can comply with both privacy laws and financial regulations simultaneously, reducing the risk of legal penalties.

\paragraph{Data Privacy in NFT Provenance Tracking}
Non-fungible tokens (NFTs) often track the provenance or ownership history of digital artworks, collectibles, and other unique assets on a blockchain. GNNs can be applied to predict future ownership transfers and analyze the relationships between buyers, sellers, and creators. However, if an owner or creator of an NFT wishes to have their transaction history or identity anonymized, GNN unlearning can be applied to remove that specific node from the model. This allows for enhanced privacy protections while maintaining the trust and transparency inherent in NFT marketplaces. Additionally, unlearning sensitive provenance data helps ensure compliance with privacy regulations that may require platforms to respect users' rights to modify or erase personal information.

\subsection{Technical Methods for Implementing GNN Unlearning in Digital Assets}
To effectively implement GNN unlearning in digital asset applications, several technical methods can be employed. The goal is to balance model accuracy, computational efficiency, and compliance with privacy regulations. Below are the leading technical approaches for GNN unlearning, particularly in the context of blockchain-based financial systems:

\paragraph{Graph Modification Techniques}
One of the most direct methods for unlearning data in a GNN is by modifying the graph itself. This involves the removal or updating of specific nodes (e.g., users, wallets, or smart contracts) and edges (e.g., transactions, relationships) that need to be forgotten. The graph modification approach recalculates the local embeddings of the impacted nodes and adjusts the learned weights associated with their interactions. Although this method ensures that the data is removed, it may require significant computational resources to update the embeddings across large, dense networks.

\paragraph{Selective Retraining via Subgraph Sampling}
Selective retraining focuses on updating only the part of the GNN that is directly affected by the unlearned data, rather than retraining the entire model from scratch. By isolating the affected subgraph (e.g., a specific transaction chain or wallet cluster), this method applies localized updates to the embeddings and weights. Subgraph sampling can minimize computational overhead while ensuring that the unlearned data no longer influences future predictions. This method is particularly well-suited to blockchain applications where the interactions between nodes are often localized, such as in specific trading pairs on a decentralized exchange.

\paragraph{Node Embedding Perturbation}
Node embedding perturbation techniques alter the learned representations of the nodes to minimize the influence of specific nodes or edges without explicitly removing them from the graph. This method injects noise into the node embeddings associated with the data that needs to be unlearned, effectively obfuscating its presence while maintaining the structural integrity of the graph. Although embedding perturbation is faster than retraining, it introduces a trade-off between privacy and model accuracy. Careful calibration is required to ensure that the perturbation does not degrade the performance of the GNN in critical applications such as fraud detection or asset management.

\paragraph{Inverse Gradient Descent}
Inverse gradient descent techniques aim to reverse the effect of the training data on the GNN’s parameters by updating the model in the opposite direction of the original training gradient. This method recalculates the gradients associated with the deleted data and applies them inversely, effectively unlearning the data without requiring full retraining. Although inverse gradient descent is computationally efficient, it requires access to the original training data and gradients, which may not always be feasible in decentralized systems where training data is distributed across nodes.

\paragraph{Federated GNN Unlearning}
Federated learning has gained popularity in decentralized systems, where models are trained across multiple nodes without sharing raw data. Federated GNN unlearning introduces a decentralized approach to forgetting, where individual nodes in the network can locally delete specific data from their part of the model. According to \cite{mcmahan2017communication}, the global GNN model is then updated using aggregated parameters from nodes that have undergone unlearning. This method ensures compliance with privacy requirements without compromising the decentralized nature of blockchain networks. Federated unlearning also minimizes the need for centralized retraining and aligns with the principles of privacy-preserving machine learning.

\subsection{Evaluation Metrics for GNN Unlearning in Digital Assets}
For GNN unlearning methods to be effectively integrated into digital asset applications, their performance must be rigorously evaluated across several dimensions. Below are the key metrics used to assess GNN unlearning techniques:

\paragraph{Unlearning Effectiveness}
This metric measures the success of removing specific data points from the model without leaving any residual traces. Unlearning effectiveness is often evaluated by testing whether the GNN’s predictions have been altered in response to the removal of the target data. For digital assets, this might involve verifying that unlearned wallets or transactions no longer influence risk assessments or market predictions.

\paragraph{Model Utility Preservation}
Maintaining the utility of the GNN after unlearning is critical, especially in financial markets where the model’s predictions directly influence decision-making. Model utility is evaluated by assessing the GNN’s performance on tasks such as transaction risk prediction, fraud detection, or token price forecasting after unlearning. Ideally, the removal of specific data should not significantly degrade the overall performance of the GNN on tasks unrelated to the unlearned data.

\paragraph{Scalability and Efficiency}
Given the size and complexity of blockchain networks, scalability is a critical metric for evaluating GNN unlearning methods. Efficient unlearning techniques should be able to handle large-scale graphs with millions of nodes and edges, minimizing both time and resource consumption. Real-time financial applications, such as decentralized exchanges and lending platforms, particularly benefit from methods that can unlearn data quickly without requiring full model retraining.

\paragraph{Privacy Guarantees and Robustness}
Privacy guarantees are essential for ensuring that unlearned data cannot be reconstructed or inferred from the remaining graph. Differential privacy and adversarial robustness measures evaluate how well the unlearning method protects against potential privacy breaches, such as adversarial attacks aimed at reconstructing deleted data. Robustness is especially important in financial systems, where adversaries might seek to exploit vulnerabilities in unlearning mechanisms to gain an unfair advantage.

\section{Conclusion and Future Direction}
Graph Neural Network (GNN) unlearning for digital assets represents a critical intersection of privacy, regulatory compliance, and performance in decentralized finance and blockchain applications. As digital assets become more pervasive, and as regulatory frameworks continue to evolve, the need for sophisticated unlearning techniques will only grow. Whether in fraud detection, risk assessment, token relationship prediction, or decentralized governance, GNNs play a pivotal role in analyzing the vast and intricate networks of relationships that underpin these systems.

\par By incorporating robust unlearning techniques, organizations can ensure that their GNN models remain compliant with privacy regulations while maintaining the high levels of performance needed for real-time financial applications. Methods such as graph modification, selective retraining, and federated GNN unlearning provide practical solutions for achieving this balance, allowing digital asset ecosystems to grow without compromising privacy or security. As GNN unlearning continues to advance, it will unlock new possibilities for data privacy in decentralized networks, ultimately shaping the future of blockchain-based financial systems.

\bibliographystyle{unsrt}  


\bibliography{references}

\end{document}